# Identifying Restaurant Features via Sentiment Analysis on Yelp Reviews

Boya Yu, Jiaxu Zhou, Yi Zhang, Yunong Cao
Center for Urban Science & Progress
New York University
New York, NY, the United States
by671@nyu.edu

*Abstract*—Many people use Yelp to find a good restaurant. Nonetheless, with only an overall rating for each restaurant, Yelp offers not enough information for independently judging its various aspects such as environment, service or flavor. In this paper, we introduced a machine learning based method to characterize such aspects for particular types of restaurants. The main approach used in this paper is to use a support vector machine (SVM) model to decipher the sentiment tendency of each review from word frequency. Word scores generated from the SVM models are further processed into a polarity index indicating the significance of each word for special types of restaurant. Customers overall tend to express more sentiment regarding service. As for the distinction between different cuisines, results that match the common sense are obtained: Japanese cuisines are usually fresh, some French cuisines are overpriced while Italian Restaurants are often famous for their pizzas.

## I. Introduction

What makes a good restaurant? What are the major concerns of customers for a great meal? Common knowledge may give general answers like delicious food, great services or pleasant environments, but they might not be true for different types of restaurants. In this paper, we are going to unveil those essential features behind all kinds of restaurants via sentiment analysis on Yelp data.

Yelp is an American multinational corporation founded in 2004 which aimed at helping people locate local business based on social networking functionally and reviews. The main purpose of Yelp is to provide a platform for customers to write review along with providing a star-rating along with an open-ended comment. Yelp data is reliable, up-to-date and has a wide coverage of all kinds of businesses. Millions of people use yelp and empirical data demonstrated that Yelp restaurant reviews affected consumers' food choice decision-making; a one-star increase led to 59% increase in revenue of independent restaurants (Luca, 2011). With the rapid growth of visitors and users, we see great potential of Yelp restaurant reviews dataset as a valuable insights repository.

As increasing amount of customers rely on Yelp for food hunting. Therefore, the review on Yelp has become an important index for food industry. In recent years there has been growing number of researches focusing on Yelp. High cited papers include a review, reputation and revenue relationship research (Luca, 2011), groupon effect (Byers, Mitzenmacher & Zervas, 2012) and an exploration of why people use Yelp (Hicks et al, 2012). Since reviews make up the greatest component for Yelp, investigations into them via machine learning techniques were expected to yield interesting discoveries. For instance, a fake review filter was developed (Mukherjee, Venkataraman & Liu, 2013) and it tested the efficiency of Yelp's abnormal spamming algorithm. This paper also applies the idea of natural language processing (NLP) to Yelp data, but it focused on the field of sentiment analysis which was conducted by a high-efficiency support vector machine (SVM) model.

Sentiment Analysis, also known as opinion mining, is the process of determining whether a text unit is positive or negative. It can have a wide range of applications such as automatically detecting feedback towards products, news and characters or improving customers' relation model.

To automate the extraction or classification of sentiment from sentiment reviews, sentiment analysis (Basant et al., 2015) uses the natural language processing (NLP), text analysis and computational techniques. Becoming one of the hot area in decision making, sentiment analysis is widely used in many fields such as Consumer information, Marketing, books, application, websites, and Social Media (Matthew et al., 2015). This analysis is divided into many levels (Thomas, 2013): document level (Ainur et al., 2010), sentence level (Noura et al., 2010), word/term level (Nikos et al., 2011) or aspect level (Haochen and Fei, 2015).

Various approaches have been used to evaluate the sentiment underneath the words and expressions or documents. Some of the most common machine learning algorithms used in NLP fields include Naive Bayes (NB), Maximum Entropy (ME), Support Vector Machine (SVM) (Joachims, 1998), and unsupervised learning (Turney, 2002). Before the rapid development of neural network based methods (Santo & Gatti, 2014) most recently, Linear SVMs often give the best performance (Mullen & Collier, 2004) in NLP.

In this paper we conducted novel analyses and discovered several interesting implications from Yelp reviews. The overall sentiment polarity showed a preference on service in the reviews, which might allude that customers 'self-select' the food they like. In the other hand, we could mine many valuable insights that cannot be directly revealed on the dashboard of websites like Yelp. Yelp's dashboard merely shows an overall rating towards a business rather than several ratings for various aspects of businesses, while we considered it at a more granular word-level.

The paper is organized as follows: Part I gives the background overview of Yelp and NLP and briefly introduces our natural language analysis on Yelp reviews. Part II describes the data and the methods for extracting information from Yelp reviews. Part III presents the results of the models. Part IV includes our discussion with discoveries within different cuisines. Part V concludes all and implies future applications in other fields.

## II. DATA AND METHODS

### A. Data Description

The dataset was originated from the online Yelp Dataset Challenge, consisting of five parts which provides us with 566,000 basic business information (e.g., hours, address, ambience), 2.2 million customer reviews as well as 519,000 tips by 552,000 users. The total size of data is about 2.39GB.

For this analysis, we focused on reviews for restaurants and used the customer reviews and business attributes data. These two datasets are both in json format. After filtering restaurants out of all business, there were 1,363,242 customer reviews collected from 77,445 different restaurants. Most of those restaurants are in Arizona, Nevada and North Carolina and the dataset includes a huge variety of cuisine types (See tables below).

| State | AZ | NV | NC | Others |
|---|---|---|---|---|
| Number of restaurants | 32615 | 21233 | 6162 | 17435 |

Figure 1 Distribution of restaurants by state

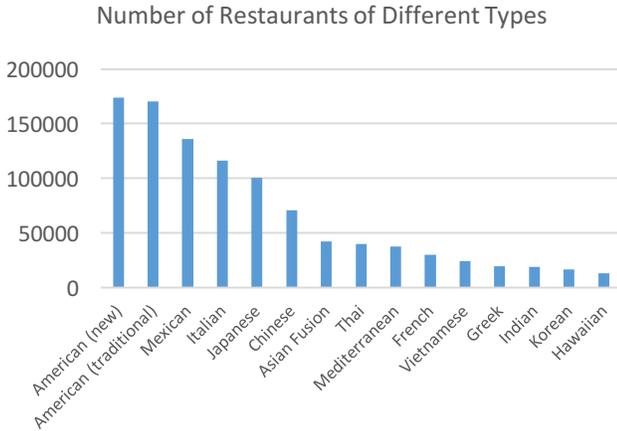

Figure 2 Distribution of reviews by cuisine type

Attributes in review data include business id, full address, price range, business categories and etc. Attributes in review data include review content, rating, business id and etc. The attributes we used were business id, business categories, review content and rating. Specifically, review content was the corpus for our analysis; rating was the identifier for discriminating positive or negative sentiment; business id served as the key for data munging and the business categories served as the key for grouping.

### B. Data Cleaning

The business dataset was merged with the reviews dataset by the attribute "business id". After that, words in each review were separated and the punctuations were removed so that a "bag of words" was generated for each review. Finally, we stemmed, lemmatized and filtered out the stop words in each bag of words using both the in-built list in Python's NLTK package.

The merged review-business data were randomly separated into training, validation and testing set according to ratio 3:2:5.

Specifically, we assumed and labeled reviews with ratings greater or equal to 4 as positive while the rest as "negative". This decision was made based on our observation of the distribution of ratings.

### C. Methods

Our methods for extracting characteristics of different types of restaurants consist of two parts. First, a Support Vector Machine (SVM) model was applied to differentiate positive and negative words in reviews, and further to get a word score to understand how positive or how negative the words were. Then the impact of scores, either negative or positive from different words within reviews of different restaurant category was analyzed.

We applied two different feature selection method for the SVM: "bag-of-words": the frequencies of various words appeared in each review and "tf-idf": the term frequency–inverse document frequency statistic. The labels were "positive" or "negative" distinguished based on the value the rating.

The tf-idf statistic we would use is:

$$\text{tfidf}(t, d) = \text{tf}(t, d) \log \frac{N}{|\{d \in D : t \in d\}|}$$

where,
tfidf$(t, d)$ is the value for term $t$ in document $d$,
tf$(t, d)$ is the frequency of term $t$ in document $d$,
$N$ is the total number of documents,
$|\{d \in D : t \in d\}|$ is the number of documents containing $t$

Pegasos Algorithm was used to solve the SVM since it was proved to have high computational efficiency when putting forward by Shwartz, Singer, Srebro and Cotter in 2011. To align with the notation used in the Pegasos paper, we're considering the following formulation of the SVM objective function.

$$\min_{w \in R^n} \frac{\lambda}{2} \|\omega\|^2 + \frac{1}{m} \sum_{i=1}^{m} \max\{0, 1 - y_i \omega^T x_i\}$$

where,
$\lambda$ is the regularization term,
$\omega$ is the vector we aim to estimate, indicating a score for each word,
$m$ is the number of samples (reviews),
$y_i$ is either 1 or -1, representing if the review is positive or negative,

$x_i$ is the feature vector of each review

Pegasos is a stochastic subgradient descent method with varied step size. The pseudocode is given below.

---
Input: $\lambda > 0$. Choose $\omega_1 = 0, t = 0$
While epoch $\leq$ max_epochs
  For $j = 1, \ldots, m$ (assumes data is randomly permuted)
    $t = t + 1$
    $\eta_t = 1/(t\lambda)$;
    If $y_j \omega_t^T x_j < 1$
      $\omega_{t+1} = (1 - \eta_t \lambda)\omega_t + \eta_t y_j x_j$
    Else
      $\omega_{t+1} = (1 - \eta_t \lambda)\omega_t$

---

Since each word was treated as an individual feature, a sparse feature matrix with very high dimensions would be generated in our model. To deal with this problem, a Python dictionary instead of a list (vector) would be set up for each review, providing information for only the words appear in the review. This avoided involving numerous zeroes in the list. Massive computation power was saved though this alternation.

At last, the scores of words to the test dataset were applied and the accuracy of our classifier was evaluated.

Both models with the bag-of-words and the tf-idf feature would be experimented with different regularization parameters. Different regularization parameters were used for features extracted through bag-of-words and tf-idf method.）We conduct experiments for two different word features, the bag-of-words which is a simple count of each word, and the tf-idf feature which includes the consideration of document frequency. Respectively, a regularization parameter is assigned for best validation accuracy, resulting in an optimization of the model. A comparison of the models would be made and the model with best test performance would be selected for implementing the next step.

In order to find specific words that were used to indicate customers' concerns for the restaurant, or by moving forward exploring the unique characteristic of each restaurant category, adjectives that simply describing the polarity of sentiment (i.e. "good", "amazing", "terrible" and etc.) were neglected. Also we assumed the rest of words could reflect the characteristic of different restaurant categories. To get an 'overall polarity score' for each word, the sentiment score of each word was multiplied by its average frequency among all reviews. And similarly, to get a 'polarity score' (a value that reflects the polarity of sentiment) towards each restaurant category, the sentiment score of each word was first multiplied by its frequency, and then normalized by the total number of reviews for the specific category of restaurants.

$$overall\_polarity\_score(t) = score(t) \times \frac{total\_frequency(t)}{total\_number\_of\_reviews}$$

$$polarity\_score(t,c) = score(t) \times \frac{total\_frequency(t,c)}{number\_of\_reviews(c)}$$

where,

$overall\_polarity\_score(t)$ is the index for measuring how essential word $t$ is among all reviews,

$score(t)$ is the word score calculated from the SVM model,

$total\_frequency(t,c)$ is the total frequency of word $t$ in all reviews,

$polarity\_score(t,c)$ is the index for measuring how essential word $t$ is among restaurants of type $c$,

$total\_frequency(t,c)$ is the total frequency of word $t$ in all reviews of type $c$ restaurants,

$number\_of\_reviews(c)$ is the total number of reviews of type $c$ restaurants.

Intuitively, since the SVM model actually calculate a total score for each review, and this score to some extent indicates how satisfied or discontented the customer is. The polarity score we calculated shows how much a word contributes to the score of all restaurants of a certain type. For example, the score of French restaurants is lowered by 0.15 in average due to 'overpriced' while is lowered by only 0.02 due to 'dirty'. Then we might claim that 'overpriced' displeased customers a lot more than 'dirty', and thus 'overpriced' is a more essential (negative) characteristic of French restaurants.

Then for each category of restaurants, the top positive and negative words are extracted. We may discover what are the special features for each type and the discrepancy of those restaurants providing great food around the world.

### III. RESULTS

Using the bag-of-words feature as well as by assessing the performance of different lambda on the validation set, we found that the result with best accuracy were achieved when lambda was set to 0.0003. The corresponding validation error was 11.035%.

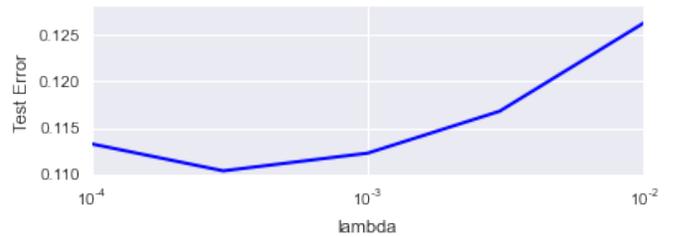

Figure 3 Test error at different regularization terms

The accuracy of the SVM classifier on the test dataset is 88.906% with lambda setting to 0.0003. For the other feature selection method 'tf-idf', none of the regularization parameter leads to an accuracy over 88%.

We also find that after stemming, lemmatizing and deleting stop words, the accuracy of our model is lower. Our guess is that some stop words may still have sentiment tendency. By ignoring these words, potential information was lost and lead to a lower accuracy. This viewpoint is supported in Saif's paper.4

For instance, it has been proven that customers tend to use the words "I/me/mine/his" more frequently when they want to express negative evaluation.

The top 5 negative/positive words affecting the polarity of reviews for all restaurants (the $overall\_polarity\_score(t)$ index) were list in the pictures below.

| Category | Top 5 Positive Words | | | | |
|---|---|---|---|---|---|
| American (New) | friendly | fresh | tasty | attentive | huge |
| American (Traditional) | friendly | fresh | tasty | attentive | huge |
| Asian Fusion | friendly | fresh | spicy | tasty | attentive |
| Chinese | friendly | fresh | tasty | spicy | authentic |
| French | friendly | fresh | wine | tasty | dessert |
| Greek | friendly | fresh | gyro | tasty | pita |
| Hawaiian | friendly | fresh | tasty | spicy | chicken |
| Indian | friendly | fresh | tasty | spicy | lamb |
| Italian | friendly | pizza | fresh | tasty | wine |
| Japanese | fresh | friendly | spicy | tasty | attentive |
| Korean | friendly | spicy | fresh | kimchi | tasty |
| Mediterranean | friendly | fresh | tasty | pita | lamb |
| Mexican | friendly | fresh | tacos | authentic | tasty |
| Southern | friendly | tasty | fresh | chicken | fried |
| Thai | spicy | friendly | curry | fresh | tasty |
| Vietnamese | fresh | friendly | pho | tasty | clean |

Figure 4 Top polarity score by restaurant type (positive)

| Category | Top 5 Negative Words | | | | |
|---|---|---|---|---|---|
| American (New) | dry | overpriced | slow | cold | rude |
| American (Traditional) | dry | slow | cold | rude | overpriced |
| Asian Fusion | overpriced | dry | slow | salty | rude |
| Chinese | dry | rude | salty | overpriced | cold |
| French | dry | overpriced | cold | slow | salty |
| Greek | dry | slow | rude | cold | overpriced |
| Hawaiian | dry | salty | rude | cold | slow |
| Indian | dry | slow | overpriced | rude | cold |
| Italian | overpriced | dry | rude | slow | cold |
| Japanese | slow | overpriced | cold | salty | rude |
| Korean | slow | overpriced | cold | rude | dry |
| Mediterranean | dry | slow | cold | overpriced | rude |
| Mexican | dry | slow | overpriced | rude | cold |
| Southern | dry | slow | cold | overpriced | salty |
| Thai | dry | slow | overpriced | salty | rude |
| Vietnamese | rude | dry | slow | cold | salty |

Figure 5 Top polarity score by restaurant type (negative)

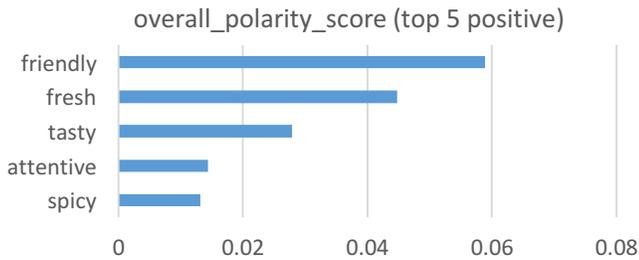

Figure 6 Top 5 Positive Words of All Types of Restaurants

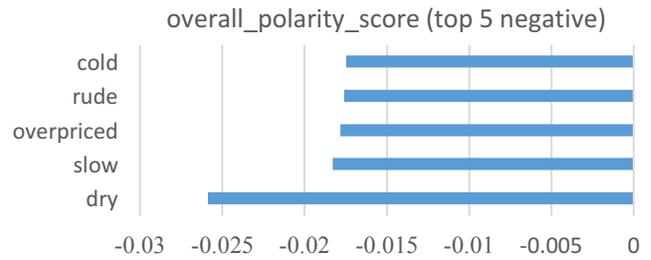

Figure 7 Top 5 Positive Words of All Types of Restaurants

Unexpectedly, the flavor of dishes does not rank first among all positive reviews. Instead, the service of restaurants seems to be the priority for most customers, since the word *friendly* ranks first and the word *attentive* is also in top 5. It could also be observed that when it comes to the flavor of food, customers value freshness more than tastiness, and spicy food seems to appeal many customers.

From the top 5 negative words list, it can be easily observed that dry food is really a taboo when it comes to the food quality, since it outnumbers others by a huge amount. In addition, we could also conclude that restaurant owners should avoid slow or rude service. As for the appearance of the word *cold*, its meaning remains to be ambiguous since we are not sure whether it refers to food or the environment of the restaurant.

The top-ranked words for different types of restaurants were also detected (See Figure 4, Figure 5), providing us a basic understanding of the characteristics for each category of restaurants.

From the negative word list, we could observe that overpriced is one of the main problems for Italian, French and eastern Asian restaurants. For Chinese and Vietnamese restaurants, the rude attitude of servers is more likely to to be the reason for a low score. On the other hand, we notice that fresh ranks first among the positive words for Japanese and Vietnamese food. There are also some dishes' names present in the positive word list, which might indicate that people prefer certain restaurants for their specific dishes, like pho in Vietnamese food and pizza in Italian food or maybe such dishes are just easier to satisfy Yelpers than others.

## IV. DISCUSSION

In this project, we have developed an efficient SVM model for discriminating positive or negative sentiment on Yelp's reviews with an accuracy of 88.906% on the test set. Other than collecting keywords from different cuisines, our model can also be used for automatically generating ratings for tips (short reviews that are not accompanied with ratings) on Yelp by assigning weights to tips using the sentiment score of words and thus giving more reasonable overall ratings for restaurants.

Based on our analysis, we found out that for most restaurant types, friendly ranks first before all the other positive comments, indicating that service might weight more than taste when people are judging a restaurant. Also for all categories of cuisines except Japanese and Vietnamese, friendly ranks the first. One possible explanation of this coincidence is that when customers decide where to have a meal, they would usually choose the specific kinds of cuisine they prefer. This kind of 'self-selection' would possibly drive the customers to focus more on the service or environment.

In addition, different characteristics are shown for different restaurant categories. Japanese and Vietnamese food received positive feedback because of freshness, while Korean and Thai restaurants received positive reviews for their spicy food. It could also be noticed that most Asian restaurants are considered to be salty including Chinese, Japanese, Southern, Thai and Vietnamese. While French, Italian, Japanese and Korean food are regarded to be overpriced, which might have something to do with their relatively better environment and service. On the contrary, waiters in Chinese and Vietnamese restaurants are mentioned to be rude, which coincides with customers' common reason, or maybe with some prejudice involved for complaining a Chinese restaurant.

Since our analysis may help to extract specific features from any set of reviews, restaurant owners can make good use of it for essential information once they received a certain amount of Yelp reviews. From those reviews they can understand why customers love or dislike their restaurants, maybe great reviews primarily due to fresh food, or perhaps unsatisfied reviews caused by too high price. Meanwhile they can also compare the restaurant with similar restaurants within the same type.

From the point of customers, a more user-satisfying recommendation and a more considerate appraisal of the restaurants can be expected if Yelp includes more features in its overall evaluation of each restaurant using the technique we have developed.

Although the performance of our model is decent, there are still a lot of spaces for improvement. One of the suggestions for future work is to try other classifiers like boosting, random forest or neural networks check whether it may outperform the SVM model. However, since the next step is to differentiate the impact of different words in varied types of restaurants, a linear-based classifier like SVM or logistic regression would be convenient. If ensemble methods like boosting and random forest or neural networks are applied for classification, the solution of this issue would be essential. For the feature selection part, variants of the tf-idf measure may be considered, or a hybrid model featuring more of a word's inherent meanings. Also, considering the large size of data, we suggest performing the work on a big data framework such as Spark or re-do the training process by more 'original' languages like C or Java to increase the computing efficiency.

## V. CONCLUSION

In this paper, we proposed an innovative method for identifying different features for restaurants of different cuisines. The method was based on a high-accuracy SVM model, calculating word scores and measuring the polarity. The essential features we discovered might not only help customers to choose their favorite cuisine, but also provide restaurants with their advantages and shortages.

On the other hand, similar procedures can be reproduced for reviews and comments in other areas like movie reviews and social media posts. Suppose someone would like to find a movie most renowned for its perfect demonstration of 'love', by operating sentiment analysis and polarity detection on IMDB or rotten tomato movie reviews, he would definitely get what he desired. That would be our anticipations in the future: gathering opinions from people, extracting information from opinions and generating suggestions from information.